\documentclass[letterpaper, 10 pt, conference]{ieeeconf}
\IEEEoverridecommandlockouts                              % This command is only needed if 
\usepackage{amsmath}
\usepackage{graphicx}
\usepackage{color}
\usepackage{multirow}
\usepackage[table,xcdraw]{xcolor}
\usepackage{float}
\usepackage{caption}
\usepackage{subcaption}
\usepackage{xspace}
\usepackage{balance}
\usepackage{makecell}
\usepackage{tabularx}
\usepackage{hyperref}
\usepackage{fancyhdr}
\usepackage{footmisc}
\usepackage{lipsum}
\usepackage{tikz}
\usepackage{float}
\usepackage[T1]{fontenc}
\usepackage[utf8]{inputenc}
\usepackage{authblk}
\usepackage{algorithm,algorithmic}

\usepackage{amsfonts}

\DeclareMathOperator*{\argmax}{argmax}

\begin{document}

\title{\LARGE \bf Spatial Relation Graph and Graph Convolutional Network for Object Goal Navigation \thanks{*Denotes equal contribution}}
% \author[]{Bond J.}
% \thanks{MI-6}
%\title{Semantic Object Navigation using Spatial Graph and Trajectory History}
\author{D. A. Sasi Kiran$^{*1}$, Kritika Anand$^{*2}$, Chaitanya Kharyal$^{*1}$, Gulshan Kumar$^1$ \\ Nandiraju Gireesh$^1$, Snehasis Banerjee$^2$, Ruddra dev Roychoudhury$^2$, Mohan Sridharan$^3$ \\ Brojeshwar Bhowmick$^2$, Madhava Krishna}
\affil[1]{Robotics Research Center, IIIT Hyderabad, India}
\affil[2]{TCS Research, Tata Consultancy Services, India}
\affil[3]{Intelligent Robotics Lab, University of Birmingham, UK}

% \author[1]{D. A. Sasi Kiran}
% \author[2]{Kritika Anand}
% \author[1]{Chaitanya Kharyal}
% \author[1]{Gulshan Kumar}
% \author[1]{Nandiraju Gireesh}
% \author[2]{Snehasis Banerjee}
% \author[2]{}
% \author[2]{Brojeshwar Bhowmick}
% \author[1]{Madhava Krishna}
% \affil[1]{Robotics Research Center, IIIT, Hyderabad, India}
% \affil[2]{TCS Research, Tata Consultancy Services, India}

\renewcommand\Authands{ and }

\markboth{IEEE Robotics and Automation Letters, 2021}
{}
%{Anand \MakeLowercase{\textit{et %al.}}:\MakeLowercase{\textit{semNav}}}

\maketitle

\begin{abstract}
%\Madhav{Wonder if we can change the title a bit, like underplay Semantic Object Nav}
This paper describes a framework for the object-goal navigation task, which requires a robot to find and move to the closest instance of a target object class from a random starting position. The framework uses a history of robot trajectories to learn a Spatial Relational Graph (SRG) and Graph Convolutional Network (GCN)-based embeddings for the likelihood of proximity of different semantically-labeled regions and the occurrence of different object classes in these regions. To locate a target object instance during evaluation, the robot uses Bayesian inference and the SRG to estimate the visible regions, and uses the learned GCN embeddings to rank visible regions and select the region to explore next. % describe region criterea
This approach is tested using the Matterport3D benchmark dataset of indoor scenes in AI Habitat, a visually realistic simulation environment, to report substantial performance improvement in comparison with state of the art baselines. 
%Additional results and supporting material are available online: \url{https://user432.github.io/objnav-srg/}
% Matterport3D (MP3D) dataset indoor scenes. The GCN utilizes the knowledge obtained from spatial relation graph and trajectory paths learnt through exploration in indoor scenes, to arrive at which region to explore next, in-order to maximise the chance of finding the target object, by their similarity in the embedding space. 
% \Madhav{utilises the knowledge to do what? And how is this different from SRG used to find out region prediction. This distinction should come clearly as we are using SRG on two counts and this should not confuse the reviewer}
%over Success and Shortest Path Length measures for a set of MP3D dataset scenes on 19 object goal categories. In addition, we propose a mathematical model which further utilises the spatial relational graph generated earlier, to predict a list of visible regions, to enable the agent to select the best one to navigate to, in order to better perform the task at hand.
\end{abstract}

%Ablation experiments shows how the spatial %common-sense graph  and trajectory path guides and %provides a bias towards a particular viewpoint, %which allows the agent to search the target object %efficiently using optimal path.

\IEEEpeerreviewmaketitle

\section{Introduction}
\label{sec:intro}
Navigation is a fundamental task performed by a service robot, e.g., in an office or a home. Navigation tasks are broadly classified into \textit{PointGoal} tasks (go to a point in space), \textit{ObjectGoal} tasks (go to a semantically distinct object instance), and \textit{AreaGoal} tasks (go to a semantically distinct area)~\cite{anderson2018evaluation}. This work focuses on \textit{ObjectGoal} navigation tasks, also called \textit{ObjectNav}. As a motivating example, consider a service robot equipped with a camera, which has been asked by a human to go to a `sink' in a home. It is difficult for the robot to perform such a task that humans perform effortlessly. The robot needs to process sensor inputs, understand its environment, and make suitable decisions to move to the target. %In this work we focus on the former tasks and handle the planning and actuation with off-the-shelf algorithms. Humans are generally very good at the task of navigation. If a person is asked to find an object in an unknown scene, his or her decision making will be based on visual cues in current and subsequent scenes. As an example, 
Specifically, to go to a `sink', the robot needs to know that a `sink' is an object usually found in a region labeled `kitchen'. It also needs to confirm its current location based on observations of \textit{relevant} objects in view, e.g., it is in the `bedroom' because it sees a `bed' nearby.%can rewrite after eg
In addition, it needs to use knowledge of the environment and the objects in its view to estimate regions that are traversable, and select the region that is most likely to lead to the target object. %the agent can infer if it is indeed in `bedroom' and get a list of the other visible regions from other objects in the view, using a novel probabilistic mathematical model we propose in this paper, which uses a spatial relational graph built using multiple trajectory walks in the scenes, to imitate common-sense and human-like understanding. 
In the current example, the robot knows it is very unlikely to find a `sink' in the `bedroom', and decides to move to the `living room', an adjacent region. %is extremely low if not zero. Hence, it chooses to explore one of the visible regions based on a critera utilising the embeddings of the target object and the visible regions, which will be explained in detail in the methodology section. Suppose the region it goes to next is living room, and now it sees different objects and  again gets a list of the visible regions, and say one of the objects is 'Oven' which enable the probabilistic mathematical model to predict 'Kitchen' as one of the visible regions. Now, since the embedding similarity of the target object Sink with this region is going to be more than the rest, it enters the kitchen, looks around and is able to locate the target `sink' which has high co-occurrence probability with `microwave oven'. This thought process though appears natural to humans is not trivial for embodiment in agents. Hence, the paper focuses on adding objective computational intelligence to the agent to carry out the `out-of-view' object finding task.  
The robot does not find a sink in the living room but it does observe an `oven' in a nearby region. It reasons that the region is most probably a `kitchen' because that is where an oven is most likely to be found. Since the robot knows that a kitchen is very likely to contain a sink, it decides to move to the corresponding region where it finds a sink.
% \begin{figure}[tb]
%   \includegraphics[width=3.3in]{images/semnav21intro2.jpg}
%   \caption{Example of semantic ObjectGoal Navigation task; service robot is asked by the human user to go to the 'sink' in a home environment.}
%   \label{fig:semnavintro}
%   \vspace{-1em}
% \end{figure}
Our framework makes the following novel contributions towards realizing this motivating scenario:
%The main contributions of the paper are as follows:\\
\begin{itemize}
    \item An approach that uses the robot's trajectories in similar environments to learn a \textit{Spatial Relational Graph} (SRG) %[Section~\ref{section:common_sense_graph_generation}] 
    that models the probability of proximity of different semantically-labeled regions to each other and the occurrence of specific object classes in each region.
    
    \item An approach that uses a \textit{Graph Convolutional Network} (GCN) operating on the historical trajectories and the learned SRG to learn
    the \textit{embeddings} %[section \ref{section:GCN}] 
    of each region and object based on their co-occurrence.
    
    \item A Bayesian inference approach %[section \ref{section:method_srg}] 
    that uses the SRG during evaluation to incrementally process the robot's current observations of specific objects and to estimate the labels of the regions visible in its current location.
    
    \item %A model for visible region prediction, which will be exploiting the Spatial-Relational graph that will be generated for training the region embeddings in another module of the proposed pipeline.
    An approach that uses the GCN-based embeddings to select the visible region to explore next, %[section \ref{section:similarity_score}], 
    computing for each region the likelihood of leading to a region with an instance of the target object class.
    
%    \item Analysis of the proposed probabilistic models and graph neural networks for visible region estimation in comparable results. Thereby, replacing computationally expensive and not-so-transparent neural network with a transparent probabilistic mathematical model which is reusing the spatial relational graphs generated for another module as input and outputting decisions that can be explained.
\end{itemize}
We use off-the-shelf algorithms for planning a path and moving a robot to a desired location and abstract away the object recognition task by assuming accurate recognition of observed objects in images of any given scene. The framework is evaluated using benchmark indoor scenes from the Matterport3D (MP3D) dataset~\cite{Matterport3D} and baseline methods in the visually realistic AI Habitat simulation environment~\cite{savva2019habitat}. A marked improvement in relevant measures in comparison with state of the art baselines is shown. Additional results and supporting material are available online: \url{https://user432.github.io/objnav-srg/}.

%\Madhav{What is this novel pipeline? What are the elements of novelty? I think before the contribution section we need a paragraph that briefs this pipeline and things novel there in the way Kritika makes use of GCN in a novel way and how SRG priors are being computed. Also you need to mention about what did it translate to in terms of results and improvements over baselines. Mention quantitative percentages.}
%\input{2_problem}
\section{Related Work}\label{sec:related_work}
We review related work on the ObjectNav task, focusing on state of the art data-driven methods.

\textbf{Mapping based approaches:}
The use of data-driven methods to learn a semantic map or an occupancy map to assist in the ObjectNav task continues to be a popular approach. These methods often use a dedicated module or a (deep) neural network, e.g., the use of a neural network to obtain a mao that is then used to sample a long-term goal to guide exploration~\cite{chaplot2020learning,ramakrishnan2022poni, chaplot2020neural}. There has also been work on using a neural network to estimate occupancy for the related PointNav task, i.e., to reach a point instead of an instance of an object class~\cite{ramakrishnan2020occupancy}. Instead of relying on a map, our framework uses a spatial relational graph and embeddings of the visible regions and objects to guide exploration. 

\textbf{End-to-End approaches:}
Data-driven methods have been developed to directly move to a given goal based on sensor inputs by learning to predict actions instead of building multiple linked components~\cite{mirowski2017learning,DBLP:journals/corr/abs-1903-01959}. This includes the use of Reinforcement Learning (RL) methods~\cite{zhu2016targetdriven, DBLP:journals/corr/abs-1805-06066}.

% \textbf{Reinforcement Learning based approaches:} Previous works such as \cite{} 

\textbf{Graph based approaches:}
Relational graphical models have been trained and used to select actions for navigation~\cite{wu2019bayesian,savinov2018semiparametric, DBLP:journals/corr/abs-2109-02066,sunderhauf2019keys}. One method builds a relational graph during training to encapsulate the relational dependencies between different regions in the scene~\cite{wu2019bayesian}. This graph is updated periodically during testing using a Convolutional Neural Network(CNN)-based region predictor network. Another method builds a topological graph during exploration, with nodes representing the locations that are used to select sub-goals~\cite{savinov2018semiparametric}. There has also been work on building a graph with region nodes, zone nodes, and object nodes, with one of the zone nodes being selected as the sub-goal that is reached using RL methods~\cite{DBLP:journals/corr/abs-2109-02066}. In another method, the graph's nodes are a few landmarks objects and robot poses, and an RL agent is trained to navigate to all possible objects~\cite{sunderhauf2019keys}. There are also methods that exploit graphical relations in different ways to aid navigation~\cite{du2020learning,qiu2020learning,DBLP:journals/corr/abs-2111-14422}. Many of the methods discussed above focus on specific simulators or datasets and make corresponding assumptions. Our framework uses a relational graph as well, but to capture the relational dependencies between both the regions and the objects during training, and make decisions during evaluation. It is also used to do region prediction based on a probabilistic model which exploits this graph.

% Some related work that uses GCN to develop the policy level navigation are \cite{sunderhauf2019keys}, \cite{vijay2019generalization}, \cite{qiulearning}. The proposed method makes an improvement on the state of art by using trajectory data priors along with a learnt region-object joint embedding learned from a GCN trainig to perform the ObjectNav task. 

% \Madhav{The current trend is to divide Related Work (RW) section into subsections and write the related work for those subsections. Finally to contrast this work with ours and say how we are different with respect to these literature. That is important. For example, we can divide the RW into say Perception Frameworks (how have people modeled the scene), Action Frameworks (how have people formulated policies, actions). If this is there in the earlier RAL paper, please do make use of that}

\begin{figure*}[!htb]
\begin{center}
  \includegraphics[width=0.9\textwidth]{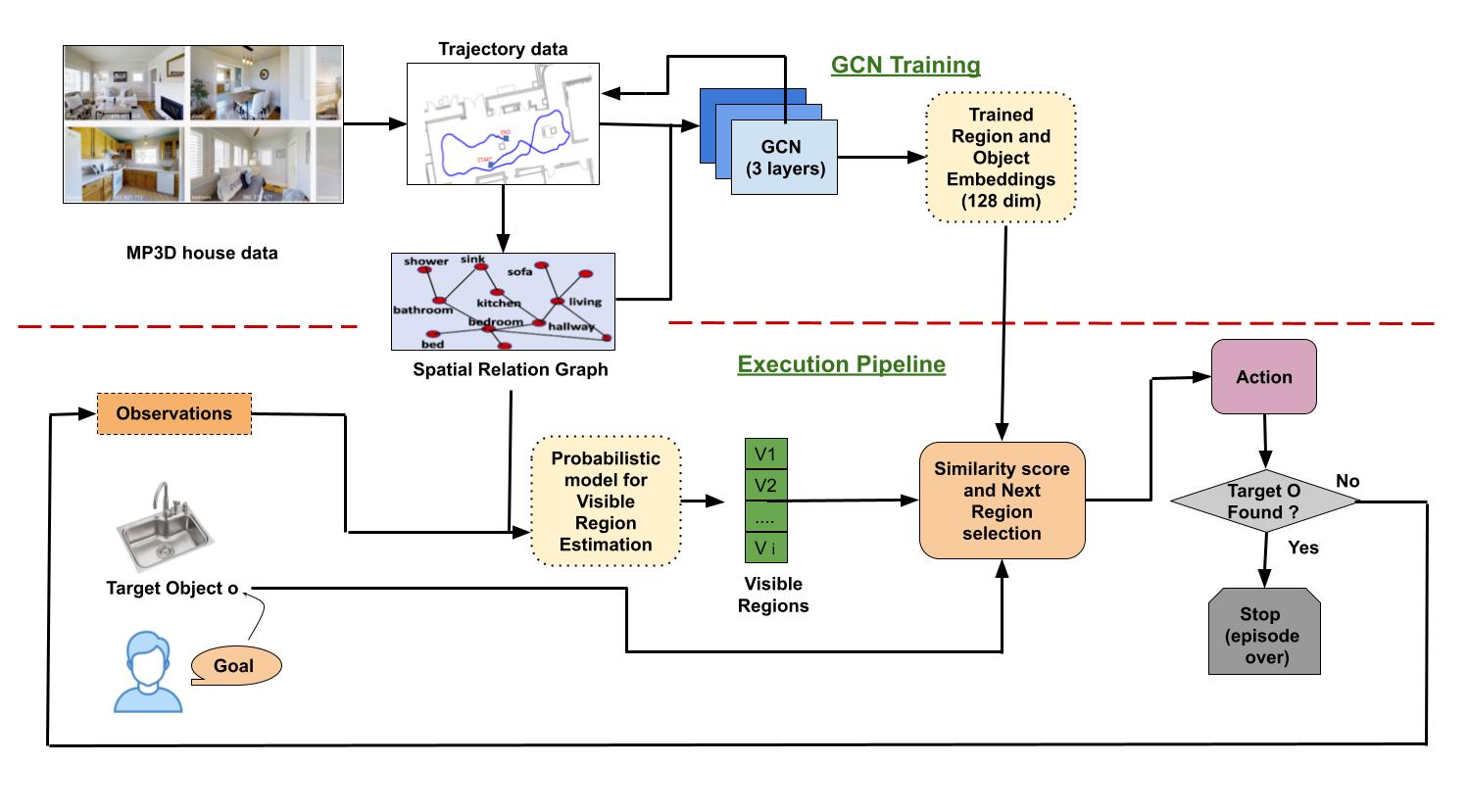}
  \vspace{-2em}
  \caption{Our ObjectNav framework has three stages each for training and evaluation. It trains and uses an SRG encoding the proximity of regions to each other and object-region co-occurrence, and a GCN-based embedding of this information and historical data of executed trajectories in indoor environments, to identify and move to the region most likely to contain an instance of the target object class in a previously unseen environment. }
  \label{fig:architecture}
  \vspace{-1em}
  \end{center}
\end{figure*}

\section{Problem Specification and Methodology}\label{sec:approach}
We focus on the \textit{ObjectNav} task in which a robot placed in a random pose in a previously unknown indoor environment is asked to find an instance of a target object class~\cite{batra2020objectnav}. 
%The robot is assumed to possess the ability to recognise objects and to plan its movement to specific locations. 
%\Mohan{needs more work to specify in terms of inputs and output} At every step, the agent has to choose the next action \textit{a} $\in$  \textit{A}, from set of actions \textit{A} = \{\textit{`Stop'}, \textit{`Move\_Forward'}, \textit{`Turn\_Left'}, \textit{`Turn\_Right'}, \textit{`Look\_Up'}, \textit{`Look\_Down'}\}. An episode is considered a `success', if the  agent is within $1.0$ metre Euclidean distance from any instance of the target object (goal).
%\Madhav{As discussed Problem Definition need not be from Habitat internals perspective. The agent needing to choose one of the 4 actions is a Habitat abstraction and not the way the problem is solved. You should mention what is the problem actually being solved.} %But this should come after related work along with the pipeline. You pose the problem, show the pipeline image and quickly delineate the solution
%\Mohan{A figure providing an overview of the entire system needs to appear here; move Figure~\ref{fig:architecture} here?}
Figure~\ref{fig:architecture} is an overview of our framework that has six stages. The first three stages, described in Sections~\ref{section:trajectory_path_generation}-~\ref{section:GCN}, correspond to the training process during which the robot executes and uses trajectories of its movement through a set of semantically-labeled scenes to compute the SRG and the GCN embeddings. The next three stages, described in Sections~\ref{section:method_srg}-~\ref{section:controller}, correspond to evaluation during which the robot uses the trained SRG and GCN to process input observations and compute a ranking of the visible regions in terms of their likelihood to lead to an instance of the target object class; an off-the-shelf planner is then used to control the robot's movement to the highest-ranked visible region. Individual stages are described below.

%The phases are: i) \textit{Generation of valid trajectory path}, ii) \textit{Generation of spatial relational graph}, iii) \textit{Generation of object and region embedding by training the GCN}, iv) \textit{Computation of Walk Score} and v) \textit{Action Controller} for Motion Planning. 

%GCN was chosen to utilize the structural information which is present in the spatial relational graph and its node features. Region and object nodes of the graph is mapped in low-dimensional embeddings in order to capture the knowledge of spatial graph and trajectory data. The successful walks included in trajectory paths while training the GCN are extracted. The main aim is to optimize the embedding structure such that regions in the trajectory path sequence have maximum possibility of co-occurrence. We make an assumption that the agent can derive successfully the region associated with the objects that are visible.

% \Madhav{What do we mean by "regions in the trajectory path sequence have maximum possibility of co-occurrence"}
% \Amarthya{By higher possibility of co-occurence, we mean that the embeddings of those regions have higher similarity, for when we have to choose which one to explore.}

\subsection{\textbf{\textit{Generating Valid Trajectories}}}
\label{section:trajectory_path_generation}
\begin{figure}[tb]
  \includegraphics[width=\linewidth]{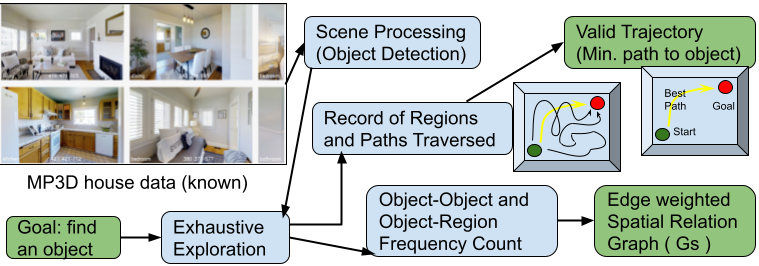}
  \caption{Generating valid trajectories by exploring target object classes in MP3D dataset scenes using AI Habitat.}
  \label{fig:trajectory}
\end{figure}

Figure~\ref{fig:trajectory} is an overview of the first step of the training process. The robot is initialized in a known \textit{MP3D} scene and given a goal to find the nearest instance of an object class. The environment is known to the robot and it moves to all the instances of the object category keeping a record of the regions encountered along the path. The path with the minimum distance to the target object is labeled as a \textit{valid trajectory} and stored. %The trajectory path consists of the regions it travelled to reach the target object. 
We generate multiple valid trajectory paths for subsequent use.

As an example, consider a trial in which the target object is a `sink' with the robot initialized in the region `bedroom'. There are instances of a sink in multiple regions of the \textit{MP3D} house/scene such as the \textit{bathroom}, \textit{laundry room}, and \textit{kitchen}; the nearest \textit{sink} to the \textit{bedroom} is in the \textit{bathroom}. If the robot instead starts in the `living room' with the same target object class (i.e., `sink'), the valid trajectory may be \{\textit{living room}, \textit{hallway}, \textit{dining room}, \textit{kitchen}\} $\xrightarrow{}$ \textit{sink}, i.e., the nearest `sink' to the `living room' is in the `kitchen'. To get to the goal target object, we find the next action using \textit{shortest path follower} algorithm, which takes into consideration the geodesic shortest path from the agent’s current position to the goal position. Overall, we obtained $18,488$ trajectory paths in AI Habitat MP3D environment. %-- these are used in path selection later in testing stage.

\begin{figure}[t]
\centering
  \includegraphics[width=0.8\linewidth]{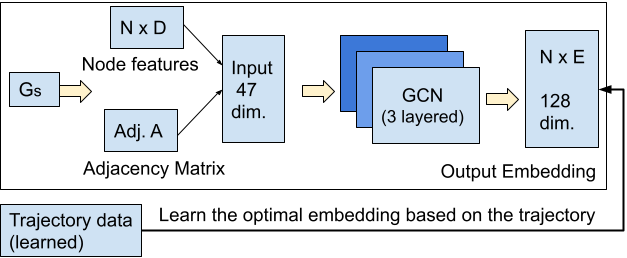}
  %\vspace{-1em}
  \caption{Training GCN to encode embedding of information in SRG based on `valid' trajectories.}
  \vspace{-1em}
\end{figure}

\subsection{\textbf{\textit{Generation of Spatial Relational Graph}}} 
\label{section:common_sense_graph_generation}
The proposed SRG graphically represents the information about spatial relations between regions and objects, which is essential for object goal navigation task. We denote this graph by \textit{G} = (\textit{V}, \textit{E}), where \textit{V} and \textit{E} represent the nodes and the edges between nodes, respectively. In particular:
\begin{itemize}
\item Each node \textit{n} $\in$ \textit{V} denotes an \textit{object} category (\textit{object node}) or the \textit{region} category (\textit{region node}); and 
\item Each edge \textit{e} $\in$ \textit{E}  denotes the relationship between \textit{region} categories or between a \textit{region} and an \textit{object} category.
\end{itemize}
We consider $2$ types of edges to encode: %which can be present in the spatial graph and scene graph :
%\begin{itemize}
%\item 
(i) `\textit{Includes}' relation between a \textit{region} and an \textit{object} category; and (ii)
%\item 
`\textit{Adjacency/proximity}' relation between a pair of regions.
%\end{itemize}

\begin{figure*}[tb]
  \includegraphics[width=\textwidth,height=5 cm]{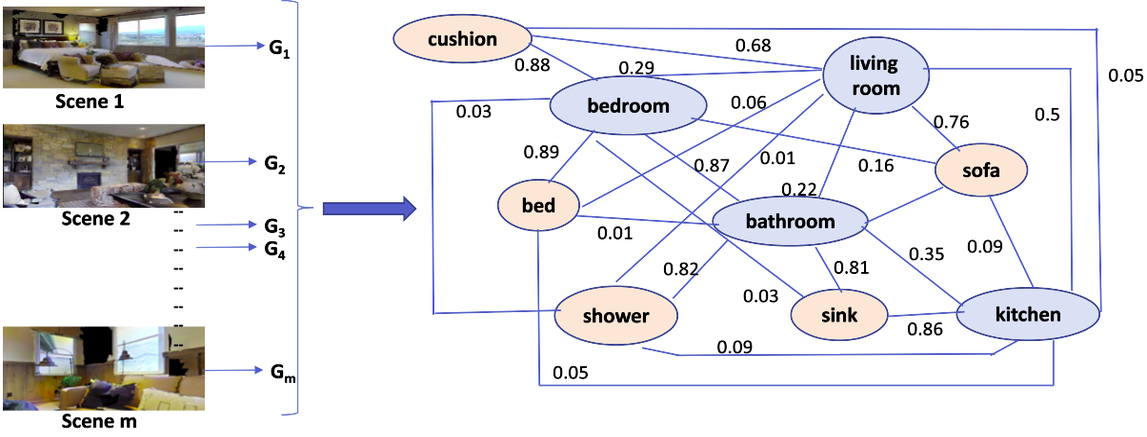}
  \vspace{-1.5em}
  \caption{Learning SRG from valid trajectories obtained through exploration of MP3D scenes. Nodes represent regions (blue) or object classes (orange), and edges encode likelihood of proximity (between regions) or occurrence of objects in regions.}
  \label{fig:common_sense_graph}
  \vspace{-1em}
\end{figure*}

In any MP3D scene with \textit{n} regions, the robot is allowed to move from region $R_{i}$ to other regions $R_j$, i.e., $i, j \in~[1,\ldots n],~i \neq j$. For each scene, an edge is created between region nodes $n_{R_k}$ and $n_{R_l}$ if nodes representing $R_{k}$ and $R_{l}$ are adjacent in the path between $R_{i}$ and $R_{j}$. Also, we create an edge between object node $n_{o}$ and a region node $n_{R_i}$, if object $o$ is in region $R_{i}$. We create such \textit{scene graphs} ($G_{1}, G_{2}, G_{3}, \ldots, G_{m} \in \mathbf{G}$) for all \textit{m} scenes in the MP3D dataset. These graphs in $\mathbf{G}$ are used to build the spatial relational graph ($G_{s}$) as depicted in Figure~\ref{fig:common_sense_graph}. 

$G_{s}$ encodes the proximity and spatial co-occurrence (frequency) statistics  of \textit{regions} and \textit{objects} extracted from the MP3D scenes explored in the valid trajectories. Specifically, the SRG associates an attribute `\textit{weight}' with the two different types of edge between the nodes; it represents the object-region co-occurrence probability for the \textit{includes} relation between a particular object and a particular region, and the proximity likelihood for the \textit{adjacency/proximity} relation between any two regions. These weights are computed for the individual graphs corresponding to the MP3D scenes ($G_{1}, G_{2}, G_{3}, \ldots, G_{m}$). 
%The `\textit{weight}' gives an estimate of the likelihood of the co-occurrence of two \textit{regions} or an inclusion of an \textit{object} within the \textit{region}. 
For example, to estimate the weight of an edge connecting object node ($n_{o=bed}$) and region node ($n_{R_i=bedroom}$) in $G_{s}$, we find the frequency of `\textit{bed} in \textit{bedroom}' in the graphs in $\mathbf{G}$ and normalize it by the frequency of `\textit{bedroom}'. The \textit{weight} for the edge between any two regions $r_{i}$ and $r_{j}$ is computed by dividing the total number of co-occurrences of $r_{i}$ adjacent to $r_{j}$ by the minimum of the frequency of the individual regions in $\mathbf{G}$.

Figure \ref{fig:common_sense_graph} shows an example SRG constructed from the scene graphs. We observe that the \textit{weight} is high ($0.89$) for `\textit{bed} in \textit{bedroom}' and  for `\textit{bedroom} adjacent to \textit{bathroom}' (\textit{weight} = $0.87$), whereas it is low ($0.05$) for `\textit{bed} in \textit{kitchen}' and `\textit{bathroom} adjacent to \textit{kitchen}' (\textit{weight} = $0.35$). These priors will help the robot discover and navigate towards the \textit{target} object from its current position. For example, seeing a \textit{dining table} from the \textit{living room} will help the robot identify and navigate to an adjacent \textit{kitchen} that is likely to contain a \textit{sink}, which is the robot's target. %which is usually present in the \textit{kitchen} and from the knowledge that \textit{kitchen} is usually adjacent to \textit{dining room} as per graph G.

\subsection{\textbf{\textit{Encoding Object-region Embeddings in GCN}}}
\label{section:GCN}
The previous section described how the SRG probabilistically encodes the proximity of regions to each other and the co-occurrence of objects and regions. For the `ObjectNav' task, it is useful to obtain a low-dimensional embedding of this information and the useful trajectories contained in the valid trajectories collected during training. %can break this long sentence

% For the `ObjectNav' task, the following priors needs to be harnessed by some means:
%     i) spatial relational graph containing spatial facts about objects which humans are expected to know in general, and 
%     ii) the trajectory data which agent can learn from the known scenes while training.
%In this context, a graph embedding representation will be useful to extract useful features from the node attributes and the structure of the spatial graph. 
Specifically, the objective is to train embeddings such that regions and objects more likely to occur together on the path to the target object have a high similarity score based on the embeddings. These embeddings are learned from the SRG and the positive trajectories by optimizing a cross-entropy loss function.
%Specifically, the objective is to train node embeddings by optimizing a cross-entropy loss function such that nodes in any valid trajectory have high similarity and high likelihood of occurring on a successful path to the target object. 
Since positive trajectories represent the path taken by the robot in a known indoor environment to successfully locate the target object, the learned embeddings are similar to the Word2vec representation for computing an embedding for words in a sentence~\cite{mikolov2013efficient} .
%In addition to capturing information provided by the  positive trajectory data created in the \textit{Phase I}, the embeddings also encodes the knowledge present in spatial relational graph created in \textit{Phase II}. 

We use a GCN to learn the embeddings because it is well-suited to capture the relationships in the SRG and trajectories. As before, we assume that the robot is able to correctly recognize objects in any observation of the current scene. 
%GCN was chosen to utilize the structural information which is present in the spatial relational graph and its node features. Region and object nodes of the graph is mapped in low-dimensional embeddings in order to capture the knowledge of spatial graph and trajectory data. The successful walks included in trajectory paths while training the GCN are extracted. The main aim is to optimize the embedding structure such that regions in the trajectory path sequence have maximum possibility of co-occurrence. We make an assumption that the agent can derive successfully the region associated with the objects that are visible.
Recall that the trained SRG $G_{s}$ has two types of edges. %, \textit{includes} and \textit{proximity}, which denote (respectively) the probability of a region containing a particular project and the probability of two regions being next to each other.
% In spatial relation graph $G_{s}$, the respective weights associated with an edge 
% 	i) between nodes $n_{R_i}$ and $n_{R_j}$ is to rate how likely the regions $R_{i}$ and $R_{j}$ co-occur together  (\textit{adjacent}/\textit{proximity}) and
%     ii) between nodes $n_o$ and $n_{R_i}$ denotes what is the likelihood of region $R_i$ having the object \textit{o} (\textit{inclusion}).
As we are only interested in the most likely links in $G_{s}$, edges %by considering the edges with high weights between the nodes (\textit{regions} and \textit{objects}). Two nodes are connected with an edge only when 
that have a \textit{weight} $\le 0.5$ are pruned.

The GCN takes two inputs during training: (i) input features for every node i, represented as a $N\times D$ matrix (N: number of nodes, D: number of input features); and (ii) graph structure in the form of an adjacency matrix A of size $N\times N$~\cite{kipf2016semi}. It produces an output of dimension $N\times E$ where $E$ is the dimension of the embedding. The \textit{region} and \textit{object} categorical values are mapped to integer values using the \textit{one-hot encoding vector} to avoid bias, i.e., the index of the node has value $1$ and other values are zeros. %This is done to avoid bias while handling categorical data (object classes). 
Specifically, a three-layer GCN takes as input the SRG in the form of an \textit{adjacency matrix} and an one-hot encoding of the features of region and object nodes. The dimension of feature vectors is the sum of the number of \textit{objects} and \textit{regions}; in this paper, we consider $19$ \text{objects} and $28$ \textit{regions}. For training, we use the graph convolutional operator (GCNConv)~\cite{kipf2016semi}; the first layer has input dimension $47$ and the last layer's output dimension is the embedding size ($128$ in this paper).

For every index $x \in \{2, 3, \cdots, n-1\}$ in a valid trajectory $\{ i_{1}, i_{2}, i_{3}, \cdots i_{n} \} \rightarrow i_{\text{target}}$, we find its prefix path $\{ i_{1}, i_{2}, i_{3}, \cdots, i_{x-1} \}$. In the loss function, we maximize the similarity of the embedding of node (i.e., region) $i_{x}$ with the embedding of $i_{n}$, and with the embedding of each node in its prefix path. Suppose the robot took the path \{\textit{living room}, \textit{hallway}, \textit{bedroom}\} $\rightarrow$ \textit{bed} to reach the nearest instance of object class \textit{bed} from its starting region.
% \{$i_{1}$, $i_{2}$, $i_{3}$ ,...., $i_{n}$\} \rightarrow $ i_{target}$ , at every index x $\in$ \{2, 3,..n-1\} in the trajectory path, we find its prefix path \{$i_{1}$, $i_{2}$, $i_{3}$ ,...., $i_{x-1}$\} . In the loss function, we maximize the embedding of $i_{x}$ with the embeddings of $i_{n}$, with each node in its prefix path. Suppose, the agent has taken the path \{ \textit{living room}, \textit{hallway}, \textit{bedroom}\} $\rightarrow$ \textit{bed} to reach the nearest target \textit{i.e} \textit{bed} from its starting region i.e \textit{living room}.
We maximize the similarity of the embedding of node \textit{hallway} with node \textit{bedroom}, and \textit{hallway} with \textit{living room}. Also, for each trajectory, we maximize the embedding of $i_{n}$ with $ i_{target}$, i.e., \textit{bedroom} with target \textit{bed} in the current example.

For each valid prefix path, we also generate invalid prefix paths in which the intermediate nodes are \textit{$i_{invalid}$} = $R$ - $i_{valid}$, where $R$ is the set of regions in the dataset and $i_{valid}$ is the set of nodes (i.e., regions) in the trajectory to the target object. For example, if the \textit{prefix path} $p_1$ is: \{\textit{living room}, \textit{hallway}, \textit{bedroom}\}, the invalid prefix path will be \{\textit{living room}, \textit{x}, \textit{bedroom}\}, where \textit{x} $\in$ $i_{invalid}$, i.e., \{\textit{living room}, \textit{bathroom}, \textit{bedroom}\}, \{\textit{living room}, \textit{dining room}, \textit{bedroom}\}, $\cdots$, \{\textit{living room}, \textit{stairs}, \textit{bedroom}\}. For an invalid prefix path $p_{invalid} =  \{i_{1}, i_{2}, i_{3}, \cdots, i_{x}\}$ and every index j $\in \{2,3,..x-1\}$, we minimize the similarity of the embedding of node $i_{j}$ with the embedding of $i_{target}$. During evaluation in a new scene, the embedding helps select a path most similar to the valid trajectories in the trained model.
 
%  \Madhav{If living room, hallway, bedroom is a valid path, will living room, drawing room, hallway, bedroom be an invalid path. This will confuse the network as on one hand we want to minimize distance between embeddings of hallway and bedroom in one case and maximize in the other case. Am I misunderstanding something here?}
%  \\
%  \Amarthya{It wont be as the intemediate regions are the ones replaced with the invalid regions. In this case, since an additional node is being added, it is not an invalid path.}

%  \begin{figure*}[h]
%   \includegraphics[width=\textwidth]{images/semNavArch.jpg}
%   \caption{Semantic Object Navigation using Spatial Relational Graph and Trajectory Data}
%   \label{fig:architecture}
% \end{figure*}

\subsection{\textbf{\textit{Visible Region Estimation using SRG}}}
\label{section:method_srg}
During evaluation (i.e., testing), the robot has to use the learned SRG and GCN embeddings to reach an instance of a target object class in a previously unseen scene. To do so, the robot first identifies the visible regions based on the observed objects and the SRG. We use Bayesian inference and some simplifying assumptions to compute the probability of a region $R$ being visible given a set of visible objects $O_v = \{o_1, o_2, \cdots o_n\}$. Without loss of generality, assume that the robot has observed two objects $o_j$ and $o_k$. Then:
\begin{align}
    p(R_i | o_j, o_k) = \frac{p(R_i , o_j, o_k)}{p(o_j, o_k)} = \frac{p(o_j|R_i, o_k)\cdot p(R_i,o_k)}{p(o_j, o_k)}
\label{eqn0}
\end{align}
where $p(R_i | o_j, o_k)$ is the probability of region $R_i$ being in the robot's view given the objects $o_j, o_k \in O_v$. %The objects $o_j$ and $o_k$ are selected from $O_v$ using KNN which will explained in detail later. 
If we make the simplifying assumption that the presence of each object is independent of the other objects, we obtain:
\begin{align}
\label{eqn1}
p(o_j|R_i, o_k) &= \frac{p(o_j, R_i, o_k)}{p(R_i, o_k)} = \frac{p(o_j, o_k|R_i) \cdot p(R_i)}{p(R_i, o_k)} \\\nonumber
    &= \frac{p(o_j|R_i) \cdot p(o_k|R_i) \cdot p(R_i)}{p(o_k|R_i)\cdot p(R_i)} \\\nonumber
    &= p(o_j|R_i)
\end{align}
which leads us to:
\begin{equation}
    p(R_i | o_j, o_k) = \frac{p(o_j|R_i) \cdot p(o_k|R_i) \cdot p(R_i)}{p(o_j, o_k)}
\label{eqn2}
\end{equation}
Since $p(o_j, o_k)$ will be a factor in the probability computation of any region, it can be treated as a constant scaling factor. We also make an assumption that all the regions are equally likely initially, leading to:
\begin{equation}
    p(R_i | o_j, o_k) = \lambda \cdot (p(o_j|R_i) \cdot p(o_k|R_i))
\label{eq:region_prob}
\end{equation}
where $\lambda$ is a constant, and we get $p(o_j|R_i) \cdot p(o_k|R_i)$ from the `includes' edges of the SRG. This computation can be performed incrementally to consider any number of observed objects in the scene, to obtain the region probabilities:
% \Madhav{Good to number the equations. Not just the last one above. Also is the very 1st equation really needed and relevant?}
% \Amarthya{Yes it is the first, which quantifies the probability for a region given a set of objects. The rest of the eqns follow from this.}
%\textit{From equation \ref{eq:region_prob}, it is evident that, this can be extended to any number of objects, as the computation reduces to a product of the conditional probabilities of the object given the region, which is being obtained from the edge weight between the object node and the region node in the spatial relational graph.}
%To get the set of visible regions, we apply the Bayes' rule described above to every visible object to get a vector $P(o_i \cdots o_k)$ of the region probabilities:
\begin{equation} 
    % \begin{bmatrix}
    % p(R_1 | o_i \cdots o_k) \\
    % p(R_2 | o_i \cdots o_k) \\
    % \vdots \\
    % p(R_n | o_i \cdots o_k)
    % \end{bmatrix}
    p(R_i | o_j, \cdots, o_k)~~\forall i\in [1, n]
\label{eqn4}
\end{equation}
While estimating the visible region list, for every visible object, we consider the set of candidate objects $\{o_j, \cdots, o_k\}$ to also include $k$ (experimentally set as 4) of its closest visible objects. This set is used to compute the region probabilities in Equation~\ref{eqn4}. The region label assigned to the visible object is that of the region with the highest probability in the vector above. Algorithm~\ref{alg:visible_regions} summarizes these steps. 
%After sufficient experimentation, we fix $K=5$.

 \begin{algorithm}[tb]
 \caption{Computing visible regions using SRG.}
 \label{alg:visible_regions}
 \begin{algorithmic}[1]
 \renewcommand{\algorithmicrequire}{\textbf{Input:}}
 \renewcommand{\algorithmicensure}{\textbf{Output:}}
 \REQUIRE SRG, $O_v = \{o_1,o_2,\cdots,o_l\}$
 \ENSURE  $V$
  \STATE visible\_regions = []
  \FOR {obj in $O_v$}
    \STATE cand\_objs = obj $\cup$~\text{nearest\_objects}(obj, $O_v$, $k$=4 )
    \vspace{5px}
    \STATE compute probabilities $p(R_i | \text{cand\_objs})~\forall i\in[1, n]$
    % \STATE $P$(\text{cand\_objs}) = $\begin{bmatrix}
    %                     p(R_1 | \text{cand\_objs}) \\
    %                     p(R_2 | \text{cand\_objs}) \\
    %                      \vdots \\
    %                     p(R_n | \text{cand\_objs})
    %                     \end{bmatrix}$
    \vspace{5px}
    \STATE visible\_regions[obj]= $\argmax_{i} \{p(R_i | \text{cand\_objs})\}$
    %   \STATE statements..
    %   \IF {($i \ne 0$)}
    %   \STATE statement..
    %   \ENDIF
  \ENDFOR \ visible\_regions 
 \end{algorithmic} 
 \end{algorithm}

% \begin{figure}[tb]
%     \centering
%     \includegraphics[width=0.5\textwidth]{images/sem_srg_4.png}
%     \caption{Segmentation of the observed scene using KNN approximation over the SRG region estimation of the object. This illustrates the performance.\Mohan{description unclear!}}
%     \label{fig:srg_seg}
% \end{figure}

\subsection{\textbf{\textit{Identifying Next Region to Explore}}}
\label{section:similarity_score}
 Among the visible regions, the robot needs to select the region to explore next. Suppose that the robot is currently able to view regions \textit{V} : \{$v_{r_1}$, $v_{r_2}$, $v_{r_3}$,...,$v_{r_l}$\}. The robot uses the trained GCN to compute the embedding of each visible region and the similarity of these embeddings with those of the \textit{target} object \textit{t}. The region with the highest similarity is chosen to be explored next.
\begin{align}
    \text{Choose} ~\argmax_{v_{r_i}}\Big(\text{\textit{Sim}}(\text{\textit{Emb}}(t), \text{\textit{Emb}}(v_{r_i}))\Big)\vert_{i\in[1,l]}
% \label{eq:region_to_explore}
\end{align}
where \textit{Sim$()$} is the cosine similarity function and \textit{Emb$()$} is the embedding output from the GCN.%, and $i \in \{ 1,2,3....,k\}$.
% Find $i$ $\in$ {1\;,2,...k} such that, \\ 
% similarity\_score = cosine\_sim(embed(\textit{t}), embed(\textit{$v_{r_k}$})) + \\ cosine\_sim(embed(\textit{t}), embed(\textit{$r_{m}$})) is maximum.
%\\
%\[ 

\subsection{\textbf{\textit{Action Controller}}}
\label{section:controller}
 After reaching the new region, if the agent sees the target object, it moves towards the object using a shortest path follower. If the target is within $1m$ Euclidean distance to the agent, the episode is terminated as a success. If the target object is not present in the new region a new set of visible regions is computed and the process is repeated until the \textit{target} is found or the maximum number of steps (350 in this work) for the episode is reached. One step corresponds to a translation movement of 0.3 m forward or backward, or a 30$^0$ rotation to the left or right. An existing off-the-shelf planner is used for planning and executing local navigation. %Figure~\ref{fig:architecture} presents the high-level overview of the proposed approach. It comprises of: a user who gives a target goal instruction like `go to sink', learned graph structure with trajectories that is used in scoring next move, action controller for robot actuation and perception (for sensing the environment like camera) and visible regions extracted from perception. 

\section{EXPERIMENTAL SETUP}
\label{sec:eval}
As baselines for comparison, we used three strategies: (i) Random action selection ; (ii) Active Neural Slam (ANS)~\cite{chaplot2020neural}; and (iii) Graph convolutional region estimator network (GCExp)~\cite{GCExp_ROMAN}. %, prior work by others in our group.
In ANS, the long term goal of the robot was chosen such that the exploration policy tried to maximize the area explored. A key component of GCExp was a region classification network (RCN), a graph neural network that mapped a Semantic graph $\mathcal{S}_{t}$ with objects from any of the $N_D$ object classes as nodes, to a probability distribution over the $N_R$ regions for each node. Its inputs include:
%, belonging to $N_R$ region classes. $\mathcal{S}_{t}$ has objects as nodes and these objects belong to any of the $N_D$ object categories. The objects are connected using K-Nearest Neighbors algorithm for constructing the graph at run-time. The input to the RCN is the semantic graph $\mathcal{S}_{t}$ in the form of
\begin{itemize}
    \item Feature vector matrix $X \in \mathbb{R}^{N \times N_D}$ for node representation, where $N$ is the number of nodes in $\mathcal{S}_{t}$. For each node $n$, the input feature vector $x_n \in \mathbb{R}^{N_D}$ is a one hot encoding of its object category.
    \item Adjacency matrix $A \in \mathbb{R}^{N \times N}$ of the graph structure. 
\end{itemize}
We experimentally evaluated the following hypotheses:
\begin{itemize}
    \item[\textbf{H1:}] The proposed framework substantially improves the success rate compared with the above baselines. %s: Random, ANS and ANS+GCExp~\cite{GCExp_ROMAN}.
    \item[\textbf{H2:}] The use of SRG for visible region estimation provides performance comparable with the use of the RCN in our framework, while significantly reducing the computational effort.
    \item[\textbf{H3:}] The SRG-based approach improves transparency in visible region estimation by explicitly identifying the objects influencing this estimation.
\end{itemize}
Hypotheses \textbf{H1} and \textbf{H2} were evaluated quantitatively while \textbf{H3} was evaluated qualitatively. 
%The output of the RCN is probability vector $p_r$ for each node $n$ in $\mathcal{S}_{t}$. We consider the number of region classes, $N_R$ as 10. Region classes that occur in very few scenes are clubbed together with those commonly occurring region classes which have similar object composition. We choose the predicted region label for a node corresponding to the highest probability value in $p_r$. We integrate the visible region estimation network RCN\cite{GCExp_ROMAN}, in our proposed pipeline, in place of the SRG based mathematical model as a baseline.
\noindent
Evaluation of hypotheses \textbf{H1}-\textbf{H2} was based on four well-established measures taken from related literature~\cite{anderson2018evaluation,GCExp_ROMAN,chaplot2020object} : 
\begin{enumerate}
    \item \textbf{Success}: ratio of the number of successful episodes to total number of episodes. An episode is successful if the robot is $\leq$ 1.0 m from the target object.

    \item \textbf{SPL} (Success weighted by path length): measures the efficiency of path taken by robot compared with optimal path; it is is computed as:
    \begin{equation*}
    SPL=\frac{1}{N} \sum_{i=1}^{N} S_i . \frac{l_i}{max(p_i, l_i)}  
    \end{equation*}
    where N is the number of test episodes, $S_i$ is a binary success indicator, $l_i$ is the length of shortest path to the closest instance of target object from the robot’s initial position, and $p_i$ is the length of path traversed by robot. 

    \item \textbf{SoftSPL}: it replaces the binary $S_i$ from SPL with a continuous success indicator $\in [0, 1]$ depending on robot's distance to the goal.
    \begin{equation*}
    SoftSPL=\frac{1}{N} \sum_{i=1}^{N} ( \underbrace{1 - \frac{d_i}{max(l_i,d_i)}}_{episode\_progress}) . (\frac{l_i}{max(p_i,l_i)})
    \end{equation*}
    where N, $l_i$, and $p_i$ are as before, and $d_i$ is the length of the shortest path to the goal from the robot’s position at episode termination.

    \item \textbf{Distance to Success (DTS)}: denotes the distance between the agent and the permissible distance to target for success at the end of an episode.
    \begin{equation*}
    DTS = max( \left \| x_T-G \right \|_2 - d, 0 )   
    \end{equation*}
    where $\left \| x_T-G \right \|_2$ is the $L2$ distance between robot and goal at the end of the episode; $d$ is the success threshold.
\end{enumerate}
As stated earlier, training and evaluation used different sets of scenes from the Matterport3D (MP3D) benchmark dataset for ObjectNav task, within the visually realistic AI Habitat simulation environment. The proposed framework is trained on the trajectories taken from 51 scenes and the testing is done on 250 episodes (each) on five MP3D scenes. 

% \begin{table*}[t]
% \caption{ Comparison of proposed method `semNav' with baselines on $19$ target objects and $6$ scenes}
% \label{table:comparison_with_baselines}
% \centering
% \begin{tabular}{p{0.3\linewidth}p{0.1\linewidth}p{0.1\linewidth}p{0.1\linewidth}p{0.1\linewidth}p{0.1\linewidth}p{0.1\linewidth}p{0.1\linewidth}}
% \hline
% Method & Success $\uparrow$ & SPL $\uparrow$ & SoftSPL $\uparrow$ & DTS $\downarrow$\\
% \hline
% Random & 0.006 &  0.0049 & 0.0363 &  6.6547 \\
% Frontier Based Exploration~\cite{yamauchi1997frontier} & 0.598 & 0.3703 & 0.3891 & 4.2478 \\
% %Frontier Based Exploration [12] + GCExp & 0.6493 & 0.3971  & 0.4130 & 3.6823 \\
% Active Neural Slam~\cite{chaplot2020learning}  & 0.6770 & 0.4954 & 0.5329 & 3.8379 \\
% %Active Neural Slam [1] + GCExp & 0.7027  & 0.5160 & 0.5507 & 3.6622 \\
% \textbf{\textbf{semNav} [this paper]} & 0.94133 & 0.658 & 0.67931 & 0.3047 \\
% \hline
% \end{tabular}
% \end{table*}

% \begin{table}[t]
% \caption{ Results on 6 object categories of SemExp \cite{chaplot2020object}}
% \label{table:comparison_with_baselines}
% \centering
% \begin{tabular}{p{0.3\linewidth}p{0.2\linewidth}p{0.1\linewidth}p{0.1\linewidth}p{0.1\linewidth}p{0.1\linewidth}{0.1\linewidth}}
% \hline
% Method & Success $\uparrow$ & SPL $\uparrow$ & DTS $\downarrow$\\
% \hline
% %semExp & 0.36 & 0.144 & 6.733\\
% \textbf{semNav} [this paper]} & 0.984 & 0.726 & 0.06\\
% \hline
% \end{tabular}
% \end{table}

% \input{5.1_experimental_setup}
% \input{5.2_baselines}
% \input{5.3_ablation_analysis}
\section{EXPERIMENTAL RESULTS}\label{sec:res}
%The proposed pipeline is evaluated over a set of scenes from the Matterport3D dataset with both, the proposed SRG based probabilistic derivation for visible regions estimation and the baseline, region estimation graph convolutional network, RCN. 
To evaluate hypotheses \textbf{H1}-\textbf{H2}, we first compared our proposed approach with the baselines in terms of the `Success' measure, with the corresponding results summarized in Table~\ref{table1}. We observe that the proposed approach performed better than the baselines, e.g., top three rows of table compared with last row. The use of SRG for visual region estimation provided performance comparable to the use of RCN for visual region estimation in our framework; there was no significant qualitative difference in the results, but incremental Bayesian inference with SRG involved much less computational effort than training and testing the substantially more complex deep network structure of RCN.

\begin{table}[tb]
\begin{center}
\caption{Comparing `Success' of proposed framework with baselines. Proposed framework provides better performance than baselines; use of SRG instead of RCN for region estimation provides comparable performance at much lower computational effort.}
\label{table1}
% \resizebox{0.8\linewidth}{!}{
\begin{tabular}{|c|c|} %change to cc for 2 columns
\hline
\multicolumn{1}{|c|}{\textbf{Method}} & \multicolumn{1}{c|}{\textbf{Success}$\uparrow$}\\
\hline
Random &  0.0056 \\
\hline
ANS & 0.69 \\
\hline
ANS+GCExp & 0.72 \\
\hline
\textbf{Framework with RCN} & \textbf{ 0.773} \\
\hline
\textbf{Framework with SRG} & \textbf{ 0.751} \\
\hline
\end{tabular}%}
\end{center}
\vspace{-1em}
\end{table}

% It is observed that the proposed mathematical model provide performance comparable with, if not better than, the neural network model. The results for this can be seen in the Tables~\ref{table1}-~\ref{table2}. %\ref{table2}, \ref{table3} and \ref{table4}. 
Next, we compared the proposed framework with the baselines using all four measures, focusing on the comparison between SRG and RCN for region estimation. The results summarized in Table~\ref{table2} indicate that SRG and RCN provide comparable performance. However, RCN involves computationally expensive training and use of a deep network for visible region estimation. SRG, on the other hand, supports incremental and efficient region estimation. Tables~\ref{table3}-\ref{table4} summarize results of a similar comparison on some representative scenes from the MP3D benchmark dataset.

\begin{table}[tb]
\begin{center}
\caption{Comparing proposed framework and baselines on all four measures, focusing on the comparison between SRG and RCN for region estimation; use of SRG provides comparable performance at much lower computational effort.}
\label{table2}
\scalebox{0.95}{
\begin{tabular}{|c|c|c|c|c|} %change to cc for 2 columns
\hline
\multicolumn{1}{|c|}{\textbf{Method}} & \multicolumn{1}{c|}{\textbf{Success} $\uparrow$} & \multicolumn{1}{c|}{\textbf{SPL} $\uparrow$} & \multicolumn{1}{c|}{\textbf{SoftSPL} $\uparrow$} & \multicolumn{1}{c|}{\textbf{DTS (m)} $\downarrow$}\\
\hline
Random &  0.0056 & 0.0032 & 0.0751 & 7.1565 \\
\hline
\textbf{Framework+RCN}  & \textbf{ 0.773} & \textbf{0.548} & \textbf{0.565} & \textbf{1.993} \\
\hline
\textbf{Framework+SRG} &  \textbf{0.751} & \textbf{0.530} & \textbf{0.553} & \textbf{2.348} \\
\hline
\end{tabular}
}
\end{center}
\vspace{-1em}
\end{table}

\begin{table}[tb]
\begin{center}
\caption{Our framework's performance on \textbf{specific scenes} in MP3D with \textbf{SRG for region estimation}.}
\label{table3} 
\begin{tabular}{|c|c|c|c|c|} %change to cc for 2 columns
\hline
\multicolumn{1}{|c|}{\textbf{Scene}} & \multicolumn{1}{c|}{\textbf{Success} $\uparrow$} & \multicolumn{1}{c|}{\textbf{SPL} $\uparrow$} & \multicolumn{1}{c|}{\textbf{SoftSPL} $\uparrow$} & \multicolumn{1}{c|}{\textbf{DTS (m)} $\downarrow$}\\
\hline
17DRP5sb8fy &  0.864 & 0.609 & 0.622 & 0.52 \\
rPc6DW4iMge &  0.700 & 0.494 & 0.520 & 2.61 \\
S9hNv5qa7GM &  0.616 & 0.393 & 0.403 & 2.67 \\
b8cTxDM8gDG &  0.868 & 0.675 & 0.689 & 1.25 \\
EDJbREhghzL &  0.708 & 0.491 & 0.529 & 4.69 \\
\hline
Average & 0.7512  & 0.530 & 0.553 & 2.348 \\
\hline
\end{tabular}
\end{center}
\vspace{-1em}
\end{table}

\begin{table}[tb]
\begin{center}
\caption{Our framework's performance on \textbf{specific scenes} in the MP3D with \textbf{RCN for region estimation}.}
\label{table4} 
\begin{tabular}{|c|c|c|c|c|} %change to cc for 2 columns
\hline
\multicolumn{1}{|c|}{\textbf{Scene}} & \multicolumn{1}{c|}{\textbf{Success} $\uparrow$} & \multicolumn{1}{c|}{\textbf{SPL} $\uparrow$} & \multicolumn{1}{c|}{\textbf{SoftSPL} $\uparrow$} & \multicolumn{1}{c|}{\textbf{DTS} (m) $\downarrow$}\\
\hline
17DRP5sb8fy &  0.878 & 0.626 & 0.639 & 0.72 \\
rPc6DW4iMge &  0.714 & 0.496 & 0.516 & 2.52 \\
S9hNv5qa7GM &  0.532 & 0.335 & 0.345 & 3.56 \\
b8cTxDM8gDG &  0.936 & 0.690 & 0.697 & 0.725 \\
EDJbREhghzL &  0.806 & 0.595 & 0.631 & 2.443 \\
\hline
Average & 0.7732  & 0.548 & 0.565 & 1.994 \\
\hline
\end{tabular}
\end{center}
\vspace{-1em}
\end{table}

%\subsection{\textbf{\textit{Ablation Study :}}}
% \begin{figure*}[tb]
% \vspace{1mm}
% \begin{tabular}{ccc}
% \includegraphics[width=0.3\textwidth,height=3.5cm]{images/trajectory_of_agent/kitchen_paper.png}\hspace{0.2cm} &
% \includegraphics[width=.3\textwidth,height=3.5cm]{images/trajectory_of_agent/hallway_paper.png}\hspace{0.2cm} &
% \includegraphics[width=.3\textwidth,height=3.5cm]{images/trajectory_of_agent/bedroom_paper.png} \\ (a) & (b) & (c) \\
% \end{tabular}
% \begin{tabular}{ccc}
% \includegraphics[width=0.3\textwidth,height=3.5cm]{images/trajectory_of_agent/bedroom_bathroom_paper.png}\hspace{0.2cm} &
% \includegraphics[width=0.3\textwidth,height=3.5cm]{images/trajectory_of_agent/bathroom2_paper.png}\hspace{0.2cm} &
% \includegraphics[width=0.3\textwidth,height=3.5cm]{images/trajectory_of_agent/bathtub_paper.png} \\ (d) & (e) & (f) \\
% \end{tabular}
% \caption{Path traversed by robot to reach the target object \textit{bathtub} starting from the (a) kitchen, and passing through (b) hallway (c) bedroom (d) door between bedroom and bathroom (e) bathroom, and reaching (f) bathtub in bathroom.}
% \label{fig:trajectory_video}
% \end{figure*}

We also explored the sensitivity of the framework's performance to the values of key parameters. Recall that we fix the radius $d$ of the space within which we consider visible objects as $10m$. We also set the maximum number $k$ of nearby objects that we considered for each object as $5$ (Section~\ref{section:method_srg}). The framework's performance for other specific values of these parameters is summarized in Tables~\ref{table5}-\ref{table6}.
\begin{table}[tb]
\begin{center}
\caption{Performance of our framework with SRG based Probabilistic Estimation Model; for d=7m.}
\label{table5} 
\begin{tabular}{|c|c|c|c|c|} %change to cc for 2 columns
\hline
\multicolumn{1}{|c|}{\textbf{Scene}} & \multicolumn{1}{c|}{\textbf{Success} $\uparrow$} & \multicolumn{1}{c|}{\textbf{SPL} $\uparrow$} & \multicolumn{1}{c|}{\textbf{SoftSPL} $\uparrow$} & \multicolumn{1}{c|}{\textbf{DTS (m)} $\downarrow$}\\
\hline
17DRP5sb8fy &  0.82 & 0.58 & 0.60 & 1.01 \\
rPc6DW4iMge &  0.65 & 0.43 & 0.45 & 3.17 \\
S9hNv5qa7GM &  0.52 & 0.338 & 0.353 & 3.58 \\
b8cTxDM8gDG &  0.77 & 0.57 & 0.59 & 2.03 \\
EDJbREhghzL &  0.54 & 0.38 & 0.43 & 6.9 \\
\hline
Average & 0.66  & 0.4596 & 0.4846 & 3.338 \\
\hline
\end{tabular}
\end{center}
\vspace{-1em}
\end{table}
\begin{table}[tb]
\begin{center}
\caption{Performance of our framework with SRG based Probabilistic Estimation Model; for k=7.}
\label{table6} 
\begin{tabular}{|c|c|c|c|c|} %change to cc for 2 columns
\hline
\multicolumn{1}{|c|}{\textbf{Scene}} & \multicolumn{1}{c|}{\textbf{Success} $\uparrow$} & \multicolumn{1}{c|}{\textbf{SPL} $\uparrow$} & \multicolumn{1}{c|}{\textbf{SoftSPL} $\uparrow$} & \multicolumn{1}{c|}{\textbf{DTS (m)} $\downarrow$}\\
\hline
17DRP5sb8fy &  0.82 & 0.60 & 0.61 & 0.87 \\
rPc6DW4iMge &  0.632 & 0.427 & 0.45 & 2.7 \\
S9hNv5qa7GM &  0.55 & 0.388 & 0.4 & 3.39 \\
b8cTxDM8gDG &  0.768 & 0.56 & 0.58 & 1.96 \\
EDJbREhghzL &  0.7 & 0.50 & 0.53 & 4.8 \\
\hline
Average & 0.694  & 0.495 & 0.514 & 2.744 \\
\hline
\end{tabular}
\end{center}
\vspace{-1em}
\end{table}

\begin{table}[tb]
\begin{center}
\caption{\textbf{Visible region estimations:} Outputs of SRG based visible region estimation model, over varied objects.}
\label{table7} 
\begin{tabular}{|c|c|} %change to cc for 2 columns
\hline
\multicolumn{1}{|c|}{\textbf{Objects in FOV}} & \multicolumn{1}{c|}{\textbf{Estimated Region}}\\
\hline
`cushion',`bed',`chair'&\\
`cabinet',`cushion' &  Bedroom \\
\hline
`towel',`shower',`sink'&\\
`towel',`chair' & Bathroom \\
\hline
'sofa', 'sofa', 'cushion'&\\
'table', 'picture' &  Living Room \\
\hline
`table',`chair',`chair'&\\
`chair',`picture' & Meeting room / Conference room\\
\hline
`counter',`cabinet',`sink'& Bar \\
\hline
`gym\_equipment',`towel',`stool'&\\
`gym\_equipment',`cabinet' & Gym / Exercise room \\
\hline
\end{tabular}
\end{center}
\vspace{-1em}
\end{table}

% \begin{table}[tb]
% \begin{center}
% \caption{Performance of our framework with SRG based Probabilistic Estimation Model; for d=7m.}
% \label{table5} 
% \begin{tabular}{|c|c|c|c|c|} %change to cc for 2 columns
% \hline
% \multicolumn{1}{|c|}{Scene} & \multicolumn{1}{|c|}{Success $\uparrow$} & \multicolumn{1}{|c|}{SPL $\uparrow$} & \multicolumn{1}{|c|}{SoftSPL $\uparrow$} & \multicolumn{1}{|c|}{DTS (m) $\downarrow$}\\
% \hline
% rPc6DW4iMge &  0.65 & 0.43 & 0.45 & 3.17 \\
% 17DRP5sb8fy &  0.82 & 0.58 & 0.60 & 1.01 \\
% S9hNv5qa7GM &  0.52 & 0.338 & 0.353 & 3.58 \\
% b8cTxDM8gDG &  0.77 & 0.57 & 0.59 & 2.03 \\
% EDJbREhghzL &  0.54 & 0.38 & 0.43 & 6.9 \\
% \hline
% Average & 0.66  & 0.4596 & 0.4846 & 3.338 \\
% \hline
% \end{tabular}
% \end{center}
% \end{table}

% \begin{table}[tb]
% \begin{center}
% \caption{Performance of our framework with SRG based Probabilistic Estimation Model; for k=7.}
% \label{table6} 
% \begin{tabular}{|c|c|c|c|c|} %change to cc for 2 columns
% \hline
% \multicolumn{1}{|c|}{Scene} & \multicolumn{1}{|c|}{Success $\uparrow$} & \multicolumn{1}{|c|}{SPL $\uparrow$} & \multicolumn{1}{|c|}{SoftSPL $\uparrow$} & \multicolumn{1}{|c|}{DTS (m) $\downarrow$}\\
% \hline
% rPc6DW4iMge &  0.632 & 0.427 & 0.45 & 2.7 \\
% 17DRP5sb8fy &  0.82 & 0.60 & 0.61 & 0.87 \\
% S9hNv5qa7GM &  0.55 & 0.388 & 0.4 & 3.39 \\
% b8cTxDM8gDG &  0.768 & 0.56 & 0.58 & 1.96 \\
% EDJbREhghzL &  0.7 & 0.50 & 0.53 & 4.8 \\
% \hline
% Average & 0.694  & 0.495 & 0.514 & 2.744 \\
% \hline
% \end{tabular}
% \end{center}
% \end{table}

\begin{figure}[tb]
    \centering
    \includegraphics[width=0.35\textwidth]{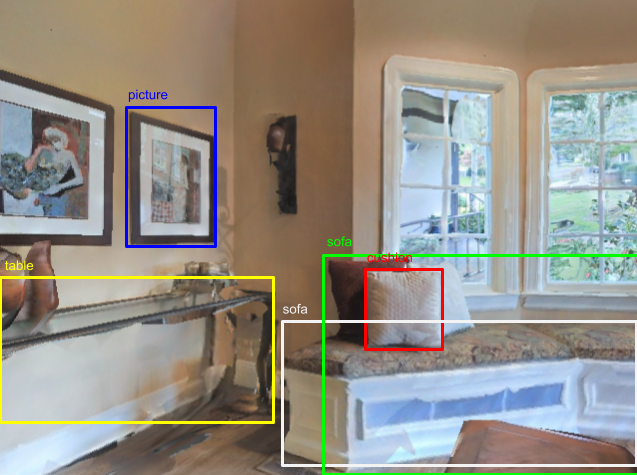}
    \caption{\textbf{Visible region estimation: Living Room}. The objects seen by the model leading to the region estimation of `Living Room' are \textit{(`sofa', `sofa', `cushion', `table', `picture')}}
    \label{fig:region_estimation}
    \vspace{-1em}
\end{figure}

Finally, to evaluate \textbf{H3}, we qualitatively compared the use of SRG-based region estimation with the use of RCN within our framework. Our framework improves transparency by providing a readily interpretable list of objects influencing the decision about specific visible regions; we also obtain a probability distribution over the candidate regions. Figure~\ref{fig:region_estimation} shows an example of region estimation on a particular image and Table~\ref{table7} shows additional examples. An added advantage of using the SRG for region estimation is the reuse of the model in very different scenes if the distribution of objects over the semantic regions is similar. %Figure~\ref{fig:trajectory_video} shows snapshots from a particular trial in which the robot starting in the \textit{kitchen} had to locate a \textit{bathtub} in the \textit{bathroom} by moving through a \textit{hallway}, \textit{bedroom}, \textit{door}, and \textit{bathroom}.

Additional results and supporting material are available online: \url{https://user432.github.io/objnav-srg/}. Since our framework is designed for the ObjectNav task, it may not provide high accuracy as a generalized navigation module due to variations in the environment and the objective functions. However, experimental results indicate that extensive training on representative and realistic scenes leads to good performance on previously unseen scenes from similar environments.

\section{Conclusion}\label{sec:conclusion}
We have described a framework for the object-goal navigation (ObjectNav) task, which requires a robot to find and move to an instance of a target object class in previously unseen scenes. The framework uses robot trajectories collected from other related scenes during a training phase to learn a Spatial Relational Graph (SRG) and Graph Convolutional Network (GCN)-based embeddings for the proximity of different semantically-labeled regions and the occurrence of different object classes in these regions. When the robot has to locate a target object instance during evaluation, Bayesian inference and the SRG are used to estimate the visible regions, and the GCN embeddings are used to rank and select the visible region to explore next. We have experimentally evaluated our framework using scenes from the Matterport3D (MP3D) benchmark dataset of indoor scenes in the visually realistic AI Habitat simulation environment. The quantitative and qualitative experimental results have demonstrated an improvement in the ability to locate the target object in comparison with baselines methods. Also, our framework significantly improves transparency while providing performance comparable with that of "black box", deep network-based approach for visible region estimation. 

Future work will extend our framework by relaxing its assumptions. First, instead of assuming that the robot can accurately recognize the observed objects in any given image, we will enable the robot to perform probabilistic object recognition, potentially by also considering neighboring objects. Second, we will extract and use knowledge about indoor regions and objects from publicly-available knowledge graphs, and explore the performance of our framework on benchmark datasets of different origin (e.g., ReplicaCAD, Gibson). Finally, we will conduct trials on a physical robot to investigate the adaptation from simulation to the real world. 

%a fundamental task for robots assisting humans we propose an approach to use region and object embeddings obtained by training GCN to solve the object goal navigation problem. The approach incorporates spatial relational knowledge of objects and regions in indoor setting, combined with valid trajectory data priors from known scenes. The efficiency of the method is proven against baseline metrics in similar settings. The ablation analysis show that the embeddings generated using spatial relations and the trajectory data together play an important role to tackle the problem. 

%\noindent
%We make an assumption that the agent knows the region associated with the visible object. Devising a method to predict the region of the visible object based on the neighbouring objects is another future direction that we will consider. In future work, we are also planning to capture common-sense and domain-specific knowledge from publicly available knowledge graphs. Another area needing attention is to carry out variations of the walk scoring function and enabling different weights for object-region and object-object edges in the spatial graph. 

%positioning of graph common sense.
%scoring mechanism.
%that does not matter..based on the viewpoint

\bibliographystyle{IEEEtran}
\bibliography{bibtex}

\end{document}